\pdfoutput=1

\documentclass[11pt]{article}

\usepackage{acl}

\usepackage{times}
\usepackage{latexsym}

\usepackage[T1]{fontenc}

\usepackage[utf8]{inputenc}

\usepackage{microtype}

\usepackage{amsmath}
\usepackage{amsfonts}
\usepackage{amssymb}
\usepackage{enumerate}
\usepackage{times}
\usepackage{latexsym}
\usepackage{graphicx}
\usepackage{algorithm}
\usepackage{longtable}
\usepackage{siunitx}

\usepackage[noend]{algpseudocode}
\usepackage{amsmath}
\usepackage{booktabs}
\DeclareMathOperator*{\argmax}{arg\,max}
\DeclareMathOperator*{\argmin}{arg\,min}

\usepackage{bm}
\usepackage{anyfontsize}
\usepackage{enumitem}
\newcommand{\red}{\color{black}}
\newcommand{\blue}{\color{black}}
\newcommand{\orange}{\color{black}}

\title{From Cognitive to Computational Modeling: Text-based Risky Decision-Making Guided by Fuzzy Trace Theory }

\author{Jaron Mar \\
  The University of Auckland \\\
  \texttt{jaron.mar@auckland.ac.nz} \\\And
  Jiamou Liu \\
  The University of Auckland \\\
  \texttt{jiamou.liu@auckland.ac.nz} \\}

\begin{document}
\maketitle
\begin{abstract}
Understanding, modelling and predicting human risky decision-making is challenging due to intrinsic individual differences and irrationality. Fuzzy trace theory (FTT) is a powerful paradigm that explains human decision-making by incorporating gists, i.e., fuzzy representations of information which capture only its quintessential meaning. 
Inspired by Broniatowski and Reyna's FTT cognitive model, we propose a computational framework which combines the effects of the underlying semantics and sentiments on text-based decision-making. In particular, we introduce Category-2-Vector to learn categorical gists and categorical sentiments, and demonstrate how our computational model can be optimised to predict risky decision-making in groups and individuals. 
\end{abstract}

\section{Introduction}
Imagine that your town is preparing for a viral outbreak 
which is projected to kill 600 people. Two alternative programs to combat the virus have been proposed. Assume that the exact scientific estimates of the consequences of the programs are {\bf Program A:} ``{\em 200 people will be saved}''; and {\bf Program B:} ``{\em 1/3 probability that all 600 lives will be saved; 2/3 probability that no lives will be saved}''. Given these choices, which program would you choose? Alternatively, if choices were presented as follows, which program would you choose?
{\bf Program C:} ``{\em 400 people will die}''; and {\bf Program D:} ``{\em 1/3 probability that no one will die and a 2/3 probability that all 600 will die}''.
This problem 
is a modified version of the {\em Asian disease problem} (ADP) \cite{tversky1981framing}, a well-studied  {\em risky decision-making problem} (RDMP) in psychology where decisions are made under risk or include probabilistic outcomes \cite{edwards1954theory}. In this RDMP, programs A and B form the {\em gain frame} where choices are worded in a positive and optimistic manner, whereas programs C and D form the {\em loss frame} where choices are written in a negative and pessimistic manner. Studies have validated that in the gain frame, humans overwhelming{\blue ly} prefer the safe choice A (72\%), whereas in the loss frame, they overwhelmingly prefer the risky choice D (78\%)  even though the choices and outcomes in both frames are equivalent \cite{tversky1981framing}. This phenomenon, known as the {\em Allais paradox}, implies that observed human choices are inconsistent with predictions based on expected utility alone, thereby confirming the influence of language, wording of choices, and sentiments on human decision-making. 



Being able to understand, model and predict human decision-making leads to many real-world applications, from predicting election results \cite{hillygus2005moral}, to improving user experience in recommender systems \cite{chen2013human}.
However, Allais paradox means that understanding the integral but complex cognitive process of decision-making, particularly in humans, is extremely challenging due to our diverse characteristics, beliefs and experiences. Furthermore, human decision-making is often fraught with irrationality
even in the presence of overwhelming evidence against some choice or beliefs \cite{simon1993decision}. This brings into question how human decision-making can be modelled with these complexities and nuances involved. This is an especially important task when considering current approaches to decision-making, such as utility theory which typically lacks any behavioural 
basis and 
ignore human sentiments 
during human decision-making \cite{lerner2015emotion}. 


Our goal is to develop a model of automated human decision-making that bridges current decision-making techniques with {\em fuzzy trace theory} (FTT), an established cognitive theory to predict group and individual decision making outlined in sections \ref{sec:task}. 
%
%
Originally proposed by Brainerd and Reyna in the 1990s, FTT 
aims to explain cognitive phenomena in memory and reasoning \cite{brainerd1990gist}. In a nutshell, FTT posits that humans form two types of mental representations, known as \textit{verbatim} which are detailed representations and \textit{gist} which are fuzzy representations that only capture the most quintessential meanings, and people prefer to make decisions based on gist rather than verbatim representations. 

In contrast with alternative cognitive and decision-making theories 
such as expected utility theory \cite{friedman1952expected} and prospect theory \cite{kahneman1980prospect}, we adopt FTT for two reasons. Firstly, FTT is the most holistic cognitive model which encompasses theories of how information is stored in memory and how memory plays an important role in our 
decision-making rather than treating decision-making as an isolated process. Because of this, FTT provides us with an extensive set of tools to explain and evaluate decision-making. Secondly, is FTTs suitability for computational modelling as conceptual parallels can be drawn between representation learning, particularly in neural-based {\blue language modelling,} and the process of creating gist representations by distilling the {\blue quintessential} information. For example, popular embedding methods for words, sentences and documents in NLP aim to create fuzzy semantic representations through dimensionality reduction of language to semantic vectors which can be viewed as gist representations of the original language \cite{liu2020representation}.

\noindent \textbf{Contributions:} 
We investigate two levels of text-based risky decision prediction tasks, group and individual-level prediction from a computational standpoint and incorporating state-of-the-art methods in NLP, we further investigate: 
\begin{itemize}[leftmargin=*]
    \item {\em How do gist representations of choices give rise to decisions?} We present a 
framework of decision-making based on gist representation learning.  

\item {\em How can we computationally encode gist representations based on the language of choices?} {\orange We outline how gist representations can be computationally encoded using techniques in NLP and propose Category-to-Vector (Cat2Vec), to learn and predict categorical embeddings of choices.}

\item {\em How can we extract the underlying sentiments of gist representations?} By extending Cat2Vec, we show how sentiments can be learnt at a categorical level; this differs from traditional approaches of sentiment analysis in NLP that examine sentiments at a text level.

\item {\em How can individual differences of individuals and groups be modelled, what impact do these differences have on decision-making?} 
We propose that individual differences are mechanisms that can encode errors at various points in the decision-making process and propose an optimisation method to infer these individual differences.

\item Finally, we demonstrate in experiments that our proposed model achieves state-of-the-art performance in predicting group and individual-based risky decision-making compared to baselines. 
\end{itemize}

\section{Task Formulation and Related Work}

Risky decision-making has been studied in many different contexts. Here we formulate \emph{$n$-choice decision-making problem} ($n$DMP): Taking as input {\blue natural language descriptions of $n$} possible choices/outcomes $\mathcal{O}$,  choose the most preferred outcome from the set of choices $\mathcal{O}$. We focus on a sub-problem known as a \emph{$n$-choice risky decision-making problem} ($n$RDMP) which is an $n$DMP where there is some risk or probabilistic outcomes associated with choices in $\mathcal{O}$, e.g., programs B and D in the ADP. Specifically, we investigate the \emph{gain-loss framing problem} which is comprised of two nRDMPs, nRDMP$_{\text{gain}}$ where choices are written as {\em gain frames} which accentuate the positive features of the text, e.g., programs A and B form a 2RDMP$_{\text{gain}}$ where `saving people's lives' is the accentuated feature. Conversely, 2RDMP$_{\text{loss}}$ where choices are written as {\em loss frames} which accentuate the negative features of the text, e.g., programs C and D form a 2RDMP$_{\text{loss}}$ where `people dying' is the accentuated feature. Additionally, choices have equivalent outcomes across both 2RDMPs. 




\subsection{Classical decision theory}
Classical decision theory abstracts the outcomes using utilities, which are numerical values that reflect desirability. For example, \textit{expected utility theory} (EUT)  
identifies the choice that maximises the expected utility assuming the axioms of rationality \cite{von2007theory}. 
However, in human decision-making these axioms are often violated, giving rise to, e.g., Allais \cite{allais1953comportement} and Ellsberg \cite{segal1987ellsberg} paradoxes. 
Generalised EUT such as {\em uncertain utility theory} \cite{gul2008measurable}, {\em cumulative prospect theory} (CPT) \cite{tversky1992advances}, and {\em multiple-criteria decision-making} (MCDM) \cite{zeleny2012multiple} were proposed to resolve these discrepancies. 
{\blue However, these classical approaches not only fail to take into account semantic information given by the working of choices which is important contextually for decision making, but they also ignore cognitive processes such as sentiments of decision-makers. }

Recent breakthroughs in NLP have led to a revolution in the breadth and robustness of problems that can be solved involving natural language by successfully capturing the underlying semantics and relationships of language. For example, neural language models 
have found resounding success in representation learning \cite{mikolov2013efficient,devlin2018bert}, the task of uncovering feature representations of language which are useful for downstream NLP tasks. One such downstream NLP task, sentiment analysis, 
has benefited largely from the application of language models such as XLNet \cite{yang2019xlnet} and ULMFiT \cite{howard2018universal}. Rapid advancements thus give hope for the development of sophisticated computational decision-making models. 


\subsection{Group/Individual-level Tasks}\label{sec:task}

In this paper, we consider two specific 2RDMP, \emph{group level risky decision making} (GL-RDM) which  the majority of psychological studies focus on and the novel task of \emph{individual level risky decision making} (IL-RDM), defined as follows. 

\noindent{\bf GL-RDM:} Given a set of observed outcomes from human RDM experiments,
each of which is described by a 5-tuple (2RDMP$_{gain}$, 2RDMP$_{loss}$, $P_{gain}$, $P_{loss}$, category), where 2RDMP$_{gain}$ is the gain frame of a 2RDMP, $P_{gain}$ is the proportion of individuals in the gain frame who chose the \textit{risky} choice, and $category$ is a grouping of similar experiments based on design and participants described in Section~\ref{exp:datasets}. 2RDMP$_{loss}$ and $P_{loss}$ can be defined similarly by replacing $gain$ with $loss$. 
 GL-RDM's objective is to predict the distribution of choice between $P_{gain}$ and $P_{loss}$ and for unseen human experiments within the same category. 


\noindent{\bf IL-RDM:} Given a set of $n$RDMPs, RDP = $\{rdp_1, rdp_2, \ldots, rdp_n\}$ where gain/loss frames of the same problem can appear as separate RDPs $rdp_i$, a set of individuals $Ivd = {ivd_1, ivd_2, \ldots, ivd_m}$ and a function which maps individuals and RDPs to their preferred choice $PC(id_i,rdp_j) = pc_{i,j}$ where $pc_{i,j}$ is individual $id_i$’s preferred choice for $rdp_j$. The objective for IL-RDM is to learn a model/mapping function for each individual which can predict an individual’s preferred choice for unseen $n$RDMPs.

\section{FTT-guided Risky Decision-making}


\noindent {\bf The BR model.} 
Broniatowski and Reyna laid out four main FTT principles in developing a cognitive model, i.e., the {\em BR model}, for the GL-RDM task \cite{broniatowski2018formal,broniatowski2014mathematical}. These principles are: ({\bf C1}) Decision choices are encoded in different levels of
gist representations, e.g., {\em categorical-}
and {\em interval-}levels based on the psychological notion of levels of measurement \cite{stevens1946theory}. ({\bf C2}) Categorical gist representations of choices are distinguished based on binary (positive/negative) sentiments and decision-makers will prefer options with positive associations. In the BR model, sentiments of categories are drawn upon social and moral principles which are stored in long-term memory, e.g., saving lives is fundamentally good. ({\bf C3}) When comparisons of categorical gist representations do not arrive at a conclusive result, the decision-maker will revert to more precise gist representations. In the BR model, gist representations compete and combine such that the simplest gist representation is chosen. ({\bf C4}) Categorical gist is encoded based on the decision-maker's prior experiences and individual differences, i.e., 
{\em need for cognition} (NFC), {\em numeracy} (NUM), and {\em risk sensitivity} (RS). 

%
Human experiments have provided evidence that the BR model is capable of explaining GL-RDM. However, being a box-arrow model, the BR model is comprised of hypothesized concepts or processes that lack precise definitions. Hence applying the model requires human interpretations and judgements on, e.g.,  notions such as gist lattices of each RDMP, the acquisition of sentiments, and individual differences. This informal nature, along with the inflexibility of the model being unable to be easily adapted to IL-RDM prevents the model from being used as an automated predictive tool. 

\smallskip

\begin{figure*}[h]
  \centering 
  \includegraphics[scale=0.55]{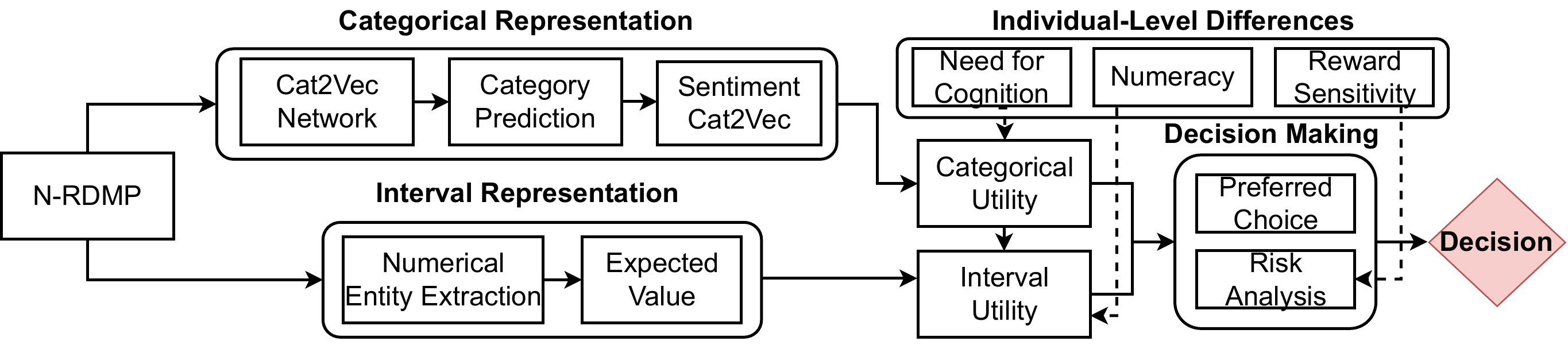}
  \caption{The main architecture of our computational model for decision making based on FTT}
 \label{fig:model}
\end{figure*}

\noindent {\bf Our  model.} Towards a fully automated tool for the RDMP tasks, we propose a computational model of 
risky decision-making that takes the input text descriptions of an RDMP and solves the GL-RDM and IL-RDM tasks automatically. The model is depicted in Figure~\ref{fig:model}. The key features of our model include: 
(1) All model components are automated, i.e., gist representations are extracted through NLP as categorical embeddings. 
(2) Categorical and interval representations are formally defined as hierarchical with the interval level encapsulating the properties and information of the categorical-level \cite{stevens1946theory}. 
(3) Individualistic differences, NFC, NUM and RS (see below) directly affect decision-making at a representational level and errors in judgement can propagate through the model adding more expressive.

\section{Computing Gist Representations}

The first challenge we tackle is the computational encoding of gist representations and how individual differences encode \textit{error}, a mechanism to model variations in human decision-making at a representation level to perform GL-RDM and IL-RDM.

\subsection{Categorical Representations}

{\em Categorisation} is the act of grouping documents into categories based on semantic or sentiment similarity. For example, the {\tt entertainment} category in the news dataset (see Section~\ref{exp:datasets}) comprises of various articles spanning multiple topics such as games, movie reviews, and celebrity gossip. The ability to categorise and recall the underlying sentiments of categories is an important prerequisite for FTT decision making asserted by principle C2 where humans prefer choices associated with categories with positive connotations over negative connotations. For example, the sentiments of the {\tt travel} category in the news dataset has a strong negative sentiment due to the news articles being collected during the outbreak of COVID-19. Given this negative sentiment, people would be dissuaded from travelling. Current sentiment analysis methods focus on granular extraction of sentiments from text rather than categories where words that are highly indicative of a category do not necessarily reveal any insights into their sentiment.

\begin{figure}[h]
\includegraphics[scale=0.31]{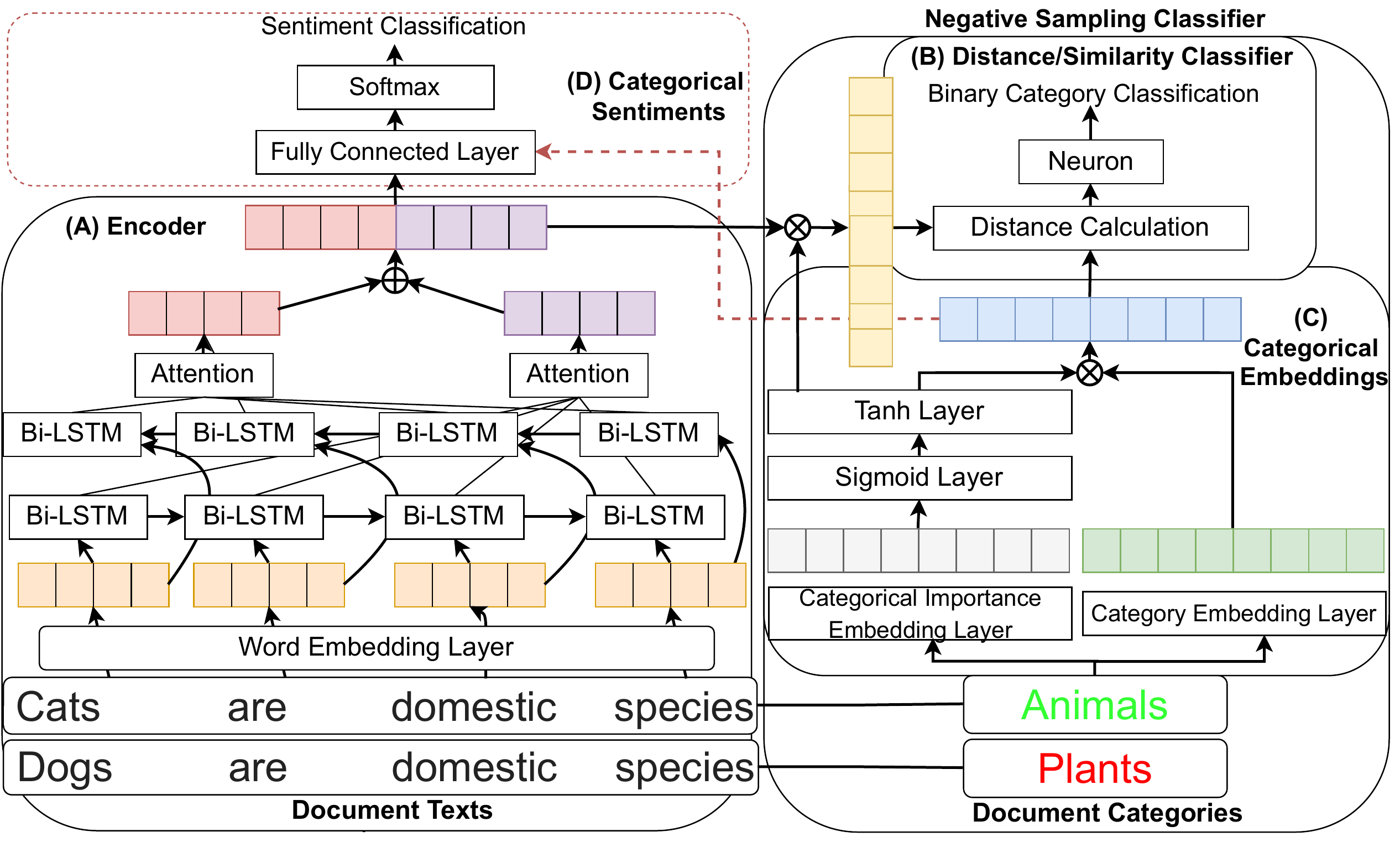}
  \caption{Cat2Vec Model}
 \label{fig:cat2vec}
\end{figure}

Vector representations of words \cite{mikolov2013efficient}, sentences \cite{devlin2018bert}, and documents \cite{le2014distributed} capture the semantic relationships between entities. At a higher level, a {\em categorical embedding} should capture semantic relations between categories of documents. To our knowledge, no such representation has been proposed. To fill this gap, we propose \textit{Category-2-Vector} (Cat2Vec) and a sentiment based extension, {\em sentiment-Cat2Vec}. 
Cat2Vec aims to find model-agnostic categorical representations that facilitates the prediction of categories from text and sentiments from categories. More formally, given a set of $M$ categories $C = \{k_1,k_2,\ldots,k_M\}$, a training set contains a number $N$ of document-category pairs $\{(d_1,c_1),\ldots, (d_N, c_N)\}$, where each $d_i$ is a document and $c_i\in C$ is the (ground truth) category of $d_i$. 
%
The objective is to maximise the average log probability $\frac{1}{N}\sum^{N}_{i=1}\log P\left(c_{i} \vert d_i\right)$
%
where $P(c_{i}\vert d_i)$ is the probability that document $d_i$ belongs to category $c_i$. {\red In sentiment-cat2vec, $P(c_{i}\vert d_i)$ is replaced by $P(s_{i}\vert c_i)$, the probability that category $c_i$ belongs to a certain sentiment class. Here, we consider only binary positive and negative sentiments.

Cat2Vec extends a contrastive learning via negative sampling 
approach by simultaneously maximising the similarity between document encodings, $v_{d_i}$, with true category embeddings, $v_{c_{i}}$, by minimising the similarity between $v_{d_i}$ and $K$ negative category embeddings defined by the objective:

\begin{multline}\label{eqn:neg_objective}
    \displaystyle \log \sigma \left((v_{c_{i}} \odot v_{i_{i}})\cdot v_{d_i}\right) \\ + \sum^{K}_{j=1} \mathbb{E}_{k_j \sim P_\text{noise}(C)}[\log \sigma ((-v_{k_j} \odot v_{i_{j}}) \cdot v_{d_i})]
\end{multline}

where $\odot$ represents element-wise multiplication, $P_\text{noise}(C)$ is a noise distribution that dictates how categories are sampled, we select a uniform distribution, $\sigma(x) = 1/(1 + \exp(-x))$ and $v_{i_{i}}$ is a category importance vector for category $i$ which is learned simultaneously with the category embedding which provides an attention-like effect over category by accentuating or diminishing certain features in the category embedding when multiplied together. Furthermore, $v_{d_i} =  Enc(d_i)$ where Enc is a document vector encoding function. In this paper we adopt a bi-directional LSTM with self-attention such that $v_{d_i} = (\overrightarrow{\alpha} \odot \overrightarrow{v}_{d_i} \| \overleftarrow{\alpha} \odot \overleftarrow{v}_{d_i})$ where $\overrightarrow{\alpha}, \overleftarrow{\alpha}$ are  self-attention weights of the forwards and backwards LSTMs, resp., and $\overrightarrow{v}_{d_i}$, $\overleftarrow{v}_{d_i}$ are the hidden state output vectors for document $d_i$ of the forward and backwards LSTMs, respectively. However, the encoder is interchangeable in Cat2Vec e.g. pretrained transformers like BERT \cite{devlin2018bert} can be used.

The novelty of our model lies in two main aspects, the introduction of a category importance vector to improve the ability of the model to learn relations between categories and the ability of our model to estimate $P(s_{i}\vert c_i)$ given labelled text documents. To estimate $P(s_{i}\vert c_i)$ we introduce an extra dense output layer (D) in figure \ref{fig:cat2vec} which predicts $P(s|d_i)$, the probability that document $i$ belongs to a certain sentiment class. In the case of binary sentiments, the joint loss becomes the binary cross-entropy loss of predicting the correct sentiment of a document plus the negative sampling loss in equation \ref{eqn:neg_objective}. After training the model we can estimate $P(s_{i}\vert c_i)$ by feeding the learned category embeddings, $(v_{c_{i}} \odot v_{i_{i}})$ into the output layers (D). Although these output layers are trained to learn $P(s|d_i)$, since the learned category embeddings are based on the document embeddings and are learned in the same semantic space, this approach gives us good estimates of $P(s_{i}\vert c_i)$.


}

Equation~\ref{eqn:CU} shows how error encoded categorical utility (CU) is calculated from categorical representations where Category(choice) is a function that takes RDM choices as inputs and outputs the underlying category related to the choice, Sentiment(category) is a function which takes a category as input and outputs the underlying sentiments related to that category as $\mathrm{pos}_{\mathrm{category}} - \mathrm{neg}_\mathrm{{category}}$. Categorical error is encoded based on NFC, an individual's tendency to engage in and enjoy cognitive activities \cite{cacioppo1996dispositional}, can introduce error at a categorical level to account for individualism. To calculate an error encoded CU, we sample from a logistic distribution which is consistent with existing literature in qualitative discrete choice models \cite{mcfadden2001economic}. Formally,  $\text{NFC} \in (0,1)$ and $CU \thicksim \mathrm{Logistic}(\mu,s)$ where $\mu = \mathbb{E}[X]$ is the expected or true utility value and $s(\text{NFC}) = |\text{NFC} - 1| \times \mathbb{E}[X]$.



\begin{equation} \small \label{eqn:CU}
    \mathrm{CU} = \mathrm{Logistic}(\mathrm{Sentiment}(\mathrm{Category}(\mathrm{choice})), \mathrm{NFC}) \\
\end{equation}

\subsection{Interval Representations}
{\em Interval representations} are a more precise representation than categorical representation. It encodes the calculation of the expected value (EV) and utility of choices. Numerical information from text can be extracted using simple text extraction or {\em named entity recognition} (NER) \cite{nadeau2007survey} where probabilities and their associated quantities can be extracted as arrays, e.g., in program B of the RDMP in the introduction the probabilities would be $[1/3, 2/3]$ and their corresponding quantities would be $[600, 0]$.  

Equation~\ref{eqn:IU} outlines the process to generate error-encoded interval utilities where CU is the categorical utility defined in equation \ref{eqn:CU} and EV is an expected value function which takes an input RDM choices and outputs the corresponding expected value associated with probabilities and quantities in choices which can extracted using standard text identification techniques such named entity recognition. Error is encoded based on NUM \cite{kahneman2003maps}, which measures a person's ability to interpret and work with numbers 
to account for individualism is calculated as $IU \thicksim \mathrm{Logistic}(\mu,s)$ where  $\text{NUM} \in (0,1)$, $\mu = \mathbb{E}[X]$ is the true expected value and $s(\text{NUM}) = Q|\text{NUM} - 1| \times \mathbb{E}[X]$ where $Q$ is number of quantities in the choice to account for error involving multiple calculations.

\begin{equation} \label{eqn:IU}
    \mathrm{IU} = \mathrm{Logisitic}(\mathrm{EV}(\mathrm{choice}), \mathrm{NUM}) \cdot \mathrm{CU}
\end{equation}


\subsection{Representations for Decision Making}\label{}

Finally, combining these representations allows us to derive the most beneficial choice in an $n$RDMP. The {\em preferred categorical}, $\text{Pref}_{\text{Cat}}$ and {\em preferred interval}, $\text{Pref}_{\text{Int}}$ choices are calculated based on which choice maximises categorical and interval utilities, respectively. 
If $\text{Pref}_{\text{Cat}}=\text{Pref}_{\text{Int}}$, there is a consensus on the best choice. If $\text{Pref}_{\text{Cat}}\neq \text{Pref}_{\text{Int}}$, there is no clear best choice. In this case, $RS$, a person's preference towards pursuing riskier but more rewarding decisions \cite{kacelnik1997risk}, is adopted as in the BR model. Risk sensitivity influences the probability of choosing the \textit{safest} or \textit{riskiest} choice in an $n$RDMP as $P(\mathrm{risky}) =1/(1 + e^{-RS})$ where $RS \in (-3, 3)$. The safest choice is one that involves the least probabilistic outcomes, whereas conversely, the riskiest choice involves the most probabilistic outcomes e.g., in the ADP in the introduction, program A is the safest as it involves one certain outcome while program B is the riskiest with two probabilistic outcomes.

\subsection{Decision Making: A Worked Example}

To demonstrate the fluidity of our model we apply our model to the ADP from the introduction.
In the gain frame, the predicted category of programs A and B using the pretrained Cat2Vec from the experiments predicts the {\tt life} category for both programs. The sentiments of the {\tt life} category predicted by Sentiment-Cat2Vec is 0.9999 positive and 0.0001 negative giving categorical utility defined as $\text{pos}_{\text{category}} - \text{neg}_{\text{category}}$ for both programs equal to 0.9998. Taking into account numerical information, the expected value of programs A and B is 200 people being saved, the interval utility is thus the expected value times the categorical utility which is 199.6 for both programs. No consensus between categorical or interval choices can be made due to unclear preferred choices for both. Thus, the final choice is decided by risk sensitivity. 

\begin{figure}[!h]

 \includegraphics[scale=0.35]{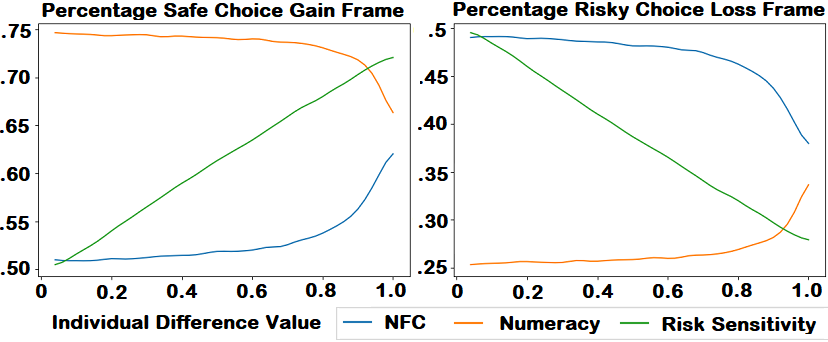}
  \caption{Effects of individual differences on the ADP.}
 \label{fig:params}

\end{figure}

Individual differences encode error and preferences into choices allowing for consensus to arise, e.g., a person with low numeracy will sample interval utilities from a logistic distribution with a larger spread than someone with high numeracy who samples utility close to the true utility. Because the error encoded utilities are sampled, the preferred choice can change on different runs of the problem; however, individual differences influence the average choice. Figure \ref{fig:params} shows a snapshot when one parameter was altered while the others were fixed and how these parameters can alter utilities to prefer certain choices across frames in the ADP. 

\section{Learning Individual Differences}

The last challenge we explore is how optimal individual-level parameters, NFC, NUM and RS can be inferred in GL-RDM and IL-RDM by optimising the following objective functions.

\begin{small}
\begin{equation}
\label{eqn:macro}
\argmin_{ivd} \sum^{\mathrm{Exp}}_{P_{\mathrm{gain}},P_{\mathrm{loss}}} |P_{\mathrm{gain}} - \hat{P}_{\mathrm{gain}}(ivd)| + |P_{\mathrm{loss}} - \hat{P}_{\mathrm{loss}}(ivd)|
\end{equation}
\end{small}

\begin{small}
\begin{equation}
\label{eqn:collective}
\begin{split}
    \hat{P}_{\mathrm{gain}}(ivd) = \sum_{exp_i \in \mathrm{Exp}} \frac{\mathbb{E}[\mathrm{CDM}(\mathrm{RDMP}_{gain}, ivd) = \mathrm{risky}]}{|\mathrm{Exp}|}
\end{split}
\end{equation}
\end{small}

\noindent {\bf GL-RDM Objective.} Is given by equation \ref{eqn:macro} where $ivd = (\mathrm{NUM}, \mathrm{NFC}, \mathrm{RS})$ are individual-level parameters which represent the characteristics of the entire group and Exp $= \{\mathrm{e}_1, \mathrm{e}_2, \ldots ,\mathrm{e}_i\}$ is the set of results based on human psychological experiments where each $\mathrm{e}$ is a 5-tuple described in the task formulation in section \ref{sec:task}.

\noindent {\bf IL-RDM Objective.} Is given by equation \ref{eqn:micro} where $ivd_i = (\mathrm{NUM}, \mathrm{NFC}, \mathrm{RS})$ is the parameters which characterises individual $i$, RDP is a set of $n$RDMPs and $PC$ is the mapping function of individuals to their preferred choices defined in the task formulation in section \ref{sec:task}. Thus, the goal is to learn optimal individual parameters for each individual which maximises the expectation that CDM chooses their true preferred choice over all RDPs.

\begin{small}
\begin{equation}
\label{eqn:micro}
\argmax_{ivd_i} \sum_j^{|\mathrm{RDP}|} \mathbb{E}[\text{CDM}(\mathrm{rdp}_j, ivd_i) = PC(ivd_i,\mathrm{rdp}_j)]
\end{equation}
\end{small}

\vspace*{-0.1cm}\section{Experiments}

\subsection{Datasets}\label{exp:datasets}

\noindent \textbf{Categorical News.} A dataset used for training/fine-tuning Cat2Vec and benchmark algorithms. The dataset contains 22601 news articles with binary sentiments labelled from various news outlets dating from February to April 2020 using the Google News API spanning 46 news categories, e.g. {\tt travel}, {\tt entertainment} and {\tt death}. 


\noindent\textbf{Group Risky Decision Making.} A dataset of 88 psychological human experiments results grouped into categories used in the evaluation of the BR model. The categories represent differences in experiment controls and participants that undertook each experiment, e.g., 'ADP; within-subjects, low PISA'. 
The category outlines the risky decision-making problem; experimental design which can be grouped into \emph{within}, where each participant is given both frames of a decision or \emph{between} subject designs, where two independent groups answer each frame; and numeracy of participants, based on the performance of the country in which the experiment took place in the Program for International Student Assessment (PISA) \cite{stacey2015international}.

\noindent \textbf{Individual 2-RDMP Prediction.} A curated dataset of 38 unique 2-RDMPs selected from various psychological experiments regarding risky decision-making answered by 121 university students using a within-subject experimental design. Of the 38 2-RDMPs, most problems contain a corresponding gain and loss frame, e.g., the ADP in the introduction, each frame is considered a separate problem. Participants selected their preferred choice from the same pre-shuffled RDMP set\footnote{See appendix, section \ref{appendix:questionnaire} for questionnaire} and no pre/post-processing of data was performed.

\subsection{Evaluation Metrics}

We apply different evaluation metrics suitable for each RDM task. For GL-RDM, we compare the true log-odds ratio (LOR) given by equation \ref{eqn:LOR}, between experimental results predicted by our model and the BR baseline model. Intuitively, the LOR measures the consistency of choices across the gain and loss frames.

\begin{equation}\label{eqn:LOR}
    \mathrm{LOR}(P_{\mathrm{gain}}, P_{\mathrm{loss}}) = \ln\left(\frac{P_{\mathrm{gain}}(1-P_{\mathrm{loss}})}{P_{\mathrm{loss}}(1-P_{\mathrm{gain}})}\right)
\end{equation}

To determine the goodness-of-fit between the predicted LOR, we apply a hypothesis test, the Wald statistic ($\chi^2$) given by equation \ref{eqn:wald}. The standard error (SE) is given by equation \eqref{eqn:se} where $n_{\mathrm{safe,gain}}$ represents the number of individuals choosing the safe choice in the gain frame, $n_{\mathrm{safe,loss}}$, $n_{\mathrm{risky,gain}}$, $n_{\mathrm{risky,loss}}$  can be derived similarly. The standard error asymptotically approaches a normal distribution when $n$ is sufficiently large; thus, the associated Wald statistic, equation \eqref{eqn:wald}, follows a chi-square distribution with one degree of freedom. 

\begin{small}
\begin{equation}\label{eqn:se}
    {\displaystyle \mathrm{SE} = \sqrt{\frac{1}{n_{\mathrm{safe,gain}}} + \frac{1}{n_{\mathrm{safe,loss}}} + \frac{1}{n_{\mathrm{risky,gain}}} + \frac{1}{n_{\mathrm{risky,loss}}}}}
\end{equation}
\end{small}

\begin{small}
\begin{equation}\label{eqn:wald}
   {\displaystyle \chi^2 = \left(\frac{\mathrm{LOR}(P_{\mathrm{gain}}, P_{\mathrm{loss}}) - \mathrm{LOR}(\hat{P}_{\mathrm{gain}}, \hat{P}_{\mathrm{loss}})}{\mathrm{SE}}\right)^2 }
\end{equation}
\end{small}

To compare the parsimony and implicitly the error between our the BR and null \cite{Busemeyer2015} models, we use the  Akaike information criterion (AIC) and Bayesian information criterion (BIC). For IL-RDM, we evaluate the accuracy of each model correctly predicting the true choices for each individual.

\subsection{Benchmark Algorithms}

In the paper, we use two different sets of baselines. For GL-RDM, we directly compare our model against the BR model. Due to the small number of experiments per grouping in the GL-RDM dataset, to maintain parity with the BR baseline, we apply the same jackknife-leave-one-out (JLOO) method  used for parameter estimation in the BR baseline model to avoid post-hoc parameter estimation \cite{busemeyer2000model}. Formally, given $m$ observed human risky decision making experiment results within a category of comparable RDPs, $\mathrm{Exp} = \{\mathrm{e}_1,\mathrm{e}_2,\ldots,\mathrm{e}_m\}$ as described in the task formulation in section \ref{sec:task}. We wish to estimate $m$ values of $G_i = (\text{NUM}_i, \text{NFC}_i, \text{RS}_i)$ where $G_i$ is group-level differences relating to observed experiments $i$ where $i = 1 \ldots m$. To achieve this, we apply the  $G_i$ can be estimated by equation \eqref{eqn:JLOO} where $\mathrm{ER}_{-i}$ is the set of experimental results excluding $\mathrm{er}_i$ as not to use the result in the estimation.

\begin{equation}\label{eqn:JLOO}
        {\displaystyle \argmin_{G_{i}} \sum^{\mathrm{ER}_{-i}}_{P_{\mathrm{gain}},P_{\mathrm{loss}}}}
        {\scriptstyle |P_{\mathrm{gain}} - \hat{P}_{\mathrm{gain}}(G_i)| + |P_{\mathrm{loss}} - \hat{P}_{\mathrm{loss}}(G_i)| }
\end{equation}

For IL-RDM, due to a lack of existing benchmark algorithms we compare our model against two baselines (1) Naive binary model using pretrained transformer language models where all RDM-choices are combined as a single input and outputs 0 or 1 corresponding to the safe or risky choice. (2) Sentiment analysis models as claim C2 asserts sentiments are highly influential in decision-making where decisions are based on choices with the highest positive sentiment. \textbf{Random:} Uniformly samples one of the available choices.\  \textbf{Vader:} A rule-based sentiment analysis for social media\cite{gilbert2014vader}. \  \textbf{XLNet:} SOTA pretrained autoregressive language model fine-tuned on the news dataset sentiments \cite{yang2019xlnet}. \  \textbf{ULMFiT:} Pretrained language model fine-tuned on the news dataset sentiments using inductive transfer learning \cite{howard2018universal}.

\subsection{Experiment Results}

\noindent {\bf GL-RDM results.} Table~1 (full table \ref{appendix:table} in Appendix) shows key discrepancies between our computational model, CDM, compared to both the actual LOR based on all 88 human experiments and those predicted by the BR model. Within each category, we find optimal group-level parameters which minimise \eqref{eqn:macro} to calculate the predicted LOR of our model for each experiment using the jackknife-leave-one-out (JLOO) method to maintain comparability between the BR model.

\begin{table}[h]

\resizebox{0.5\textwidth}{!}{
\begin{tabular}{l | c | c | c c c} \label{table:macroresults}
Reference &  Actual & BR & CDM & SE &  $\chi^2$\\
\hline 
\multicolumn{6}{c}{Standard ADP; one presentation, between-subjects, low PISA} \\
(1) \citet{tversky1981framing} & \bf{2.20} & \bf{1.65*} & \textbf{1.86} & \textbf{.26} & \textbf{1.83} \\








 
(2) \citet{mayhorn2002decisions}, Young & 2.98 & 1.68 & 1.50 & .58 & 5.89* \\
 



 \multicolumn{6}{r}{TOTAL of 14 predicted: 13 (93\%)} \\












\hline 
\multicolumn{6}{c}{Standard ADP; within-subjects, low PISA} \\
(3) \citet{LeBoeuf2003} Exp \#2 & \bf{.57} & \bf{1.05*} & \bf{.81} & \bf{.17} & \bf{.52} \\
\multicolumn{6}{r}{TOTAL of 3 predicted: 3 (100\%)} \\
\hline 
\multicolumn{6}{c}{Standard ADP; multiple presentations, between-subjects, low PISA,} \\
(4) \citet{Jou1996} & \bf{2.01} &  \bf{.87*} & \bf{1.13} & \bf{.33} & \bf{6.99*} \\
\multicolumn{6}{r}{TOTAL of 6 predicted: 5 (83\%)} \\
    \hline 
 \multicolumn{6}{c}{Other problems; multiple presentations, between-subjects, high PISA} \\
(5)  \citet{kuhberger1995framing} Plant \#2 & \textbf{2.34} & \textbf{.7*} & \textbf{.34} & \textbf{.73} & \textbf{7.50*}\\
 \multicolumn{6}{r}{TOTAL of 4 predicted: 3 (75\%)} \\
     \hline 
 \multicolumn{6}{c}{Zero-complement problems; multiple presentations, between-subjects} \\
(6)  \citet{kuhberger2010risky} Crops & -.43 & 0 & .34 & .30 & 6.83* \\
(7)  \citet{kuhberger2010risky} Fish disease & \bf{.83} & \bf{0*} & \bf{.37} & \bf{.30} & \bf{15.74*} \\
  \multicolumn{6}{r}{TOTAL of 7 predicted: 5 (71.4\%)} \\
     \hline 
 \multicolumn{6}{c}{\small“400 not saved” certain-option problems; multiple presentations, between-subjects, high
PISA} \\
(8) \citet{kuhberger1995framing} Plant \#1 &  \textbf{.49} & \textbf{-.88*}  & \textbf{.11} & \textbf{.58} & \textbf{.50} \\
(9) \citet{kuhberger1995framing} Cancer \#1 &  -1.36 & -.74 & .11 & .60 & 6.21* \\
\multicolumn{6}{r}{TOTAL of 5 predicted: 6 (83.3\%)} \\
  \hline 
\multicolumn{6}{r}{OVERALL TOTAL of 88 predicted: 82 (93.2\%)} \\
\hline 
\multicolumn{6}{l}{* Indicates results with Wald statistics over 1 degree of freedom}\\
\multicolumn{6}{l}{Note: Bolded rows indicate results where CDM outperforms or is equal to the BR model}
\end{tabular} 
}
\caption{Group Level Experiment Results.}
\end{table}

Critically, our results show that our computational model is capable of automating the prediction of human risky decision making on a wide range of RDMPs by predicting 82 of 88 (93.2\%) experiments based on the Wald statistic. These results hold even when RDMPs were manipulated to capture a wider gamut of decision making through variations on framing and truncation of choices where options were removed \cite{reyna2014developmental}. These results are comparable to carefully crafted human conducted analysis using the BR model which also predicted 82 of 88 experiments.

To further demonstrate the parsimony of our model compared to the BR and null models by applying the AIC and BIC metrics under a null CDM model where parameters are set to 0, we get AIC=14941 and BIC=14950. Whereas under the null model of the BR model, AIC=14981 and BIC=14986. In the best cases, our model outperforms all variations of the BR baseline model, with our model attaining AIC=13374 and BIC=13383 compared to AIC=13409 and BIC=13510. {\orange Furthermore, taking into consideration the relative likelihood ratio (RLR) to compare models using the AIC scores, $\exp((13374 - 13409)/2)=2.5 \times 10^{-8}$, yields a significant result where the BR baseline model is only $2.5 \times 10^{-8}$ times as probable as our model to minimize the information loss.} Thus, our model attains better goodness-of-fit compared to the BR model while using significantly fewer parameters, 3 compared to up to 172 in the BR model.



\noindent {\bf IL-RDM results.} Table~\ref{tab:micro} displays the average 5-fold cross-validation result predicting all 121 individuals' decisions for all 38 questions. Our model with a modest 63.19\% accuracy outperforms all benchmark algorithms which hover around 50\% for sentiment and 60\% for pretrained language model baselines. This reinforces that IL-RDM is a more challenging problem and although sentiment analysis is important for decision-making, current SOTA sentiment analysis is not suitable for IL-RDM and only performs comparably to random choice. It is worth noting that while pretrained language models can be naively applied to IL-RDM with competitive results, they can not be naively applied to GL-RDM which requires the simultaneous predictions of two distributions of choices across frames where often the same RDM and choices is used across all experiments within a category.

Also displayed in the table are results when using transformers as encoders within Cat2Vec and results from a minor ablation study. For IL-RDM, transformers do not significantly improve accuracy as the resulting predicted categories and sentiments of categories from RDM-choices are highly similar between encoders. In the ablation study where choices are derived based on preferred choices at different levels of representation, i.e., $\text{CDM}_{\text{Categorical}}$ and $\text{CDM}_{\text{Interval}}$, reinforces the full expressiveness of our model comes from the consensus between levels of representation and influence of individual differences.

\begin{table}[htp]
\centering
\small\setlength\tabcolsep{3pt}
\begin{tabular}{|c|c|c|c|}
\hline 
Algorithm & Accuracy $(\%)$ & Min $(\%)$ & Max $(\%)$\\ 
\hline 
Random & $ 49.87 \pm 0.4 $ & 48.99  & 51.23 \\
Vader & $51.37 \pm 0.9 $ & 48.64 & 54.33  \\
XLNet &  $49.83 \pm 2.1$  & 44.21 & 56.31\\
ULMFiT  & $47.08 \pm 2.1$ & 39.31 & 53.30\\
\hline
BERT& $61.94 \pm 0.7$ & \textbf{60.80} & 64.70  \\
GTP2 & $53.65 \pm 1.1$ & 51.44 & 57.95  \\
roBERTa & $60.83 \pm 1.7$ & 57.23 & 65.87 \\
\hline
CDM (Bi-LSTM) & $62.47 \pm 0.6$ & 60.43 & 64.11 \\
CDM (BERT) & $62.30 \pm 1.2$ & 59.54 & 65.87 \\
CDM (GTP2) & $\textbf{63.19}$ $\bf{\pm}$ $\textbf{1.2}$ & 60.64 & \textbf{68.00}  \\
CDM (roBERTa) & $62.68 \pm 1.2$ & 59.43 & 66.35  \\
\hline
Sentiment Cat2Vec & $ 53.72 \pm 1.4 $ & 48.04  & 56.67 \\
$\text{CDM}_{\text{Categorical}}$ & $ 51.86 \pm 1.3 $ & 46.38 & 54.33 \\
$\text{CDM}_{\text{Interval}}$ & $ 53.17 \pm 1.2 $  & 48.40 & 55.84  \\
\hline 
\end{tabular} 
\caption{Individual-Level Experiment Results}
\label{tab:micro}
\end{table}

\subsection{Error Analysis and Discussion}

To understand the shortfall of our model for both GL-RDM and IL-RDM, we analyse cases in which our model fails to predict human decision-making. In GL-RDM, of the 6 experiments that our model did not successfully predict, 3 of these ((4),(5) and (7) in table 1) were not predicted by the BR baseline model indicating problems with parameter estimation using JLOO as these experiments are outliers with relatively significant differences in LORs within their respective categories.

\begin{figure}[!h]
\includegraphics[scale=0.29]{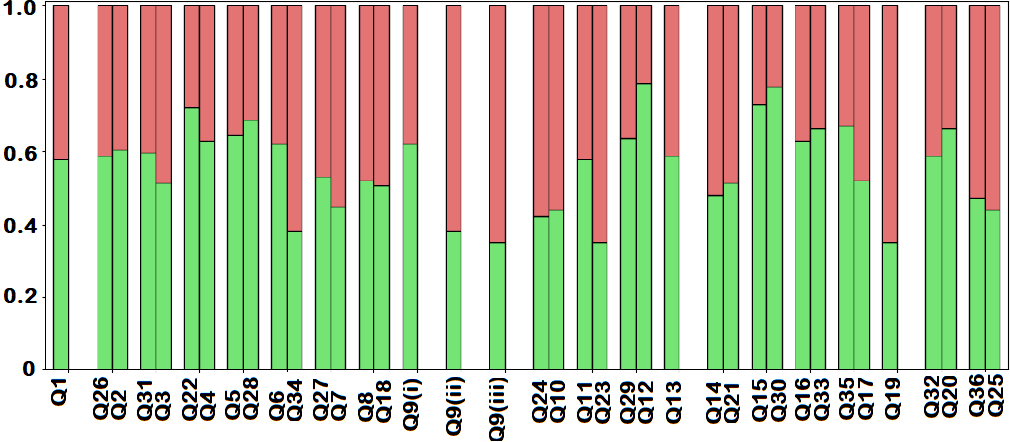}
  \caption{Ratio of choices for all questions with gain and loss frames grouped.}
 \label{fig:IL-questions}
\end{figure}

In IL-RDM, inconsistencies exist across 2-RDMPs due in part to the within-subject design as participants may notice the underlying problem causing them to compare between problems rather than independently \cite{kahneman2002representativeness}. For example, figure \ref{fig:IL-questions} shows loss frames where individuals overwhelming preferred the safe choice, e.g., Q2, Q4 and most gain-loss pairs do not show a clear distinction between safe and risky choices in opposing frames, e.g. (Q26, Q2). Both cases are inconsistent with psychological studies. 

\begin{figure}[!h]

\includegraphics[scale=0.26]{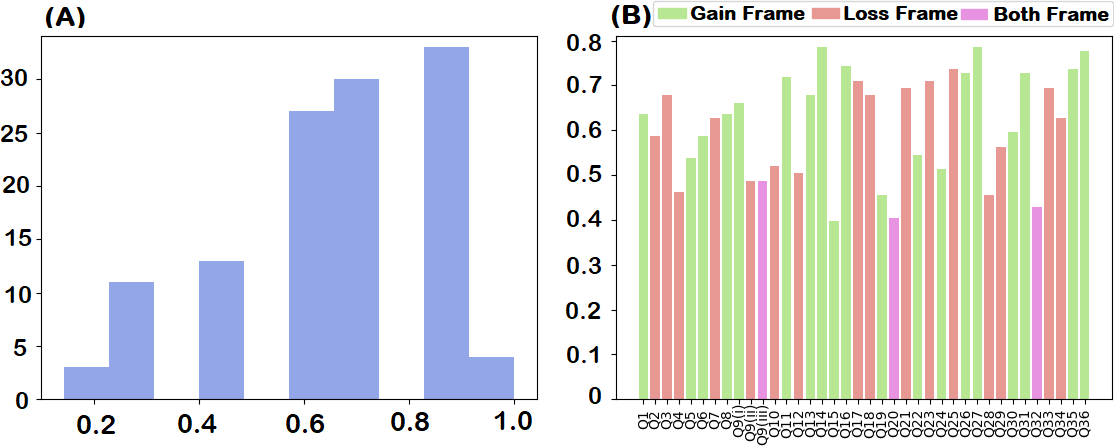}
  \caption{(A) Histogram of individual accuracy on one fold and (B) combined k-fold test accuracy per  question}
 \label{fig:IL-distribution}

\end{figure}

Quantity of data is also an issue. While the number of participants and questions answered were quite large for psychology experiments, this dataset is relatively small for machine learning tasks. Figure \ref{fig:IL-distribution}(A) shows the histogram of all 121 individuals' test accuracy on one fold. Overall, our model can predict most individual's choices accurately, but the average is lowered by some individuals our model cannot predict due to inconsistencies mentioned in (1). Also, since the size of each fold is relatively small, containing 7-8 test RDPs, any RDP not predicted correctly will cause a large decrease in accuracy. Figure \ref{fig:IL-distribution}(B) shows the percentage of correct choices our model predicts for all individuals from the combined 5-fold test questions. On average, our model predicts gain and loss RDMPs relatively equally with accuracy of $65.73\%$ and $62.84\%$ respectively. However, RDMPs with \emph{"both frame"}, contains choices with combined gain and loss wording, cannot be predicted by our model due to this duality with an average of $42.73\%$.

\section{Conclusion and Future Work}
This paper provides the first steps into a fully computational framework of risky decision-making, which adopts the cognitive and psychological basis of FTT with our model outperforming baselines in individual and group RDP prediction. Potential applications of our model are wide-ranging for scenarios in which predicting and understanding the characteristics of human risky decision-making is pivotal, e.g., the design of safety mechanisms based on how people make decisions in risky scenarios or in improving personalised recommendation systems based on understanding the users' personal traits and how they make decisions. Future work, therefore, involves adapting our model towards real-world applications, exploration of generalised decision-making and the design and evaluation of sophisticated end-to-end machine learning models for text-based decision-making. 

%
 

\bibliography{custom}
\bibliographystyle{acl_natbib}

\appendix

\section{Appendix}
\label{sec:appendix}

\subsection{Final Decision Algorithm}

Algorithm \ref{alg:decision} corresponds to the algorithm mentioned in section \ref{exp:datasets} of the main paper.

\begin{algorithm}

\caption{Computational Decision Making}\label{alg:decision}
\small
 \textbf{Input: $n$RDMP, NUM, NFC, RS} \\
 \textbf{Output: Decision/Preferred Choice}
\begin{algorithmic}[1]
    \For {choice in n-RDMP}
        \If {CU($\text{Pref}_{\text{Cat}}, \text{NFC}) <$ CU(choice, NFC)}
            \State $\text{Pref}_{\text{Cat}}$ = choice
        \EndIf
        \If {IU($\text{Pref}_{\text{Int}}, \text{NUM}) <$ IU(choice, NUM)}
            \State  $\text{Pref}_{\text{Int}}$ = choice
        \EndIf
    \EndFor 
    
    \If {$\text{Pref}_{\text{Cat}} = \text{Pref}_{\text{Int}}$}
        \State \textbf{return} $\text{Pref}_{\text{Cat}}$
    \ElsIf {Uniform(0, 1) $\leq$ RiskSensitivity(RS)}
        \State \textbf{return} Riskiest Choice
    \Else   
        \State \textbf{return} Safest Choice
    \EndIf

\end{algorithmic}

\end{algorithm}

\subsection{Evaluation Metric Calculations}

The calculations for the second type of evaluation metric we use to compare the parsimony of our model against baseline algorithms are the Akaike information criterion (AIC) and Bayesian information criterion (BIC) are given by equations \eqref{eqn:aic} and \eqref{eqn:bic} using the log-likehood calculated by equations \eqref{eqn:likelihood} and \eqref{eqn:likelihood2}. In these equations  $n_{1,1}$ is the number of people who chose the first choice (safe choice) in the first problem (gain frame), $p_{1,1}$ is the predicted proportion of subjects who chose the first choice (safe choice) in the first problem (gain frame), etc. For the AIC, $k$ is the total number of parameters of our model, 3 which correspond to each individual difference and in BIC, $n$ is the total number of data points, 176 to represent the gain and loss frames in the 88 human experiments.

\begin{multline}
    \ln[L(y_i)] = n_{1,1} \ln{p_{1,1}} + n_{1,2} \ln{p_{1,12}}\\ 
     + n_{2,1} \ln{p_{2,1}} + n_{2,2} \ln{p_{2,2}} \label{eqn:likelihood}
\end{multline}

\begin{equation}
    \ln[L(y)] = \sum_i{\ln[L(y_i)]} \label{eqn:likelihood2}
\end{equation}
\begin{equation}
    \mathrm{AIC} = 2k-2\ln[L(y)] \label{eqn:aic}
\end{equation}
\begin{equation}
    \mathrm{BIC} = k\ln(n)-2\ln[L(y)] \label{eqn:bic}
\end{equation}

To compare models using AIC, the relative likelihood ratio (RLR) given in equation \ref{eqn:relativelikelihood} can be applied which compares the probability that the BR baseline model minimises the estimated information loss compared to our CDM model given that AIC$_{\mathrm{CDM}} \leq$ AIC$_{\mathrm{BR}}$ where AIC$_{\mathrm{BR}}$ and AIC$_{\mathrm{CDM}}$ are corresponding AIC scores of each model.

\begin{equation}\label{eqn:relativelikelihood}
    \mathrm{RLR} = \exp(\frac{\mathrm{AIC}_{\mathrm{CDM}} - \mathrm{AIC}_{\mathrm{BR}}}{2})
\end{equation}

\subsection{Individual Level Questionnaire}\label{appendix:questionnaire}

Full inventory of all 36 questions used in the Individual 2-RDMP Prediction dataset:

\noindent Q1: Which of the following options do you prefer? 
    \begin{enumerate}[label=(\alph*),nosep,leftmargin=3\parindent] 
        \vspace{0.1cm} 
        \item A sure win of \$30 
        \vspace{0.1cm} 
        \item 80\% chance to win \$45 
    \end{enumerate}
\vspace{0.3cm}
Q2: Imagine that 6000 pieces of precious paintings in a world-famous museum are accidentally exposed to a disastrous chemical pollution. Two alternative plans to rescue these art treasures have been proposed. Assume that the exact estimates of the consequences of the plans made by scientists are as follows: 
    \begin{enumerate}[label=(\alph*),nosep,leftmargin=3\parindent] 
        \vspace{0.1cm} 
        \item If plan A is adopted, 4000 pieces will be destroyed by the chemical pollution. 
        \vspace{0.1cm} 
        \item If plan B is adopted, there is a one-third probability that none of these paintings will be destroyed, and two-thirds probability that all 6000 of these paintings will be destroyed. 
    \end{enumerate}
\vspace{0.3cm}
Q3: A large car manufacturer has recently been hit with a number of economic difficulties and it appears as if three plants need to be closed and 6000 employees laid off. The vice-president of production has been exploring alternative ways to avoid this crisis and has developed two plans: 
    \begin{enumerate}[label=(\alph*),nosep,leftmargin=3\parindent] 
        \vspace{0.1cm} 
        \item Plan C: This plan will result in the loss of 2 plants and 4000 jobs. 
        \vspace{0.1cm} 
        \item Plan D: This plan has a 2/3 probability of resulting in the loss of 3 plants and all 6000 jobs, but has a 1/3 probability of losing no plants and no jobs 
    \end{enumerate}
\vspace{0.3cm}
Q4: Imagine you recieve a letter from the president of a subsidiary describing a dilemma concerning whether to fight an impending patent violation suit or settle out of court that reads: If we do not agree to this proposal, PMG will file their suit. Going to court would involve the possibility of losing \$1,100,000 in damages and losing the Duraplast line. If we win in court, we will incur a small sum for legal expenses. Our corporate lawyer, Mr. Bell, and our outside law firm estimate that we have a 2 in 3 chance of losing the case. 
    \begin{enumerate}[label=(\alph*),nosep,leftmargin=3\parindent] 
        \vspace{0.1cm} 
        \item Agree to the proposal (no lawsuit) 
        \vspace{0.1cm} 
        \item Disagree to the proposal: 2/3 chance of losing the lawsuit and incurring costs of \$1100000 
    \end{enumerate}
\vspace{0.3cm}
Q5: Imagine that you have lung cancer and you must choose between two therapies: surgery and radiation. Surgery for lung cancer involves an operation on the lungs. Most patients are in the hospital for two or three weeks and have some pain around their incisions; they spend a month or so recuperating at home. After that, they generally feel fine. Radiation therapy for lung cancer involves the use of radiation to kill the tumor and requires coming to the hospital about four times a week for six weeks. Each treatment takes a few minutes and during the treatment, patients lie on a table as if they were having an x-ray. During the course of the treatment, some patients develop nausea and vomiting, but by the end of the six weeks they also generally feel fine. Thus, after the initial six or so weeks, patients treated with either surgery or radiation therapy feel about the same. 
    \begin{enumerate}[label=(\alph*),nosep,leftmargin=3\parindent] 
        \vspace{0.1cm} 
        \item Surgery: Of 100 people having surgery, 90 live through the postoperative period, 68 are alive at the end of one year and 34 are alive at the end of five years. 
        \vspace{0.1cm} 
        \item Radiation Therapy: Of 100 people having radiation therapy, all live through treatment, 77 are alive at the end of one year and 22 are alive at the end of five years. 
    \end{enumerate}
\vspace{0.3cm}
Q6: Imagine that you brought \$6000 worth of stock from a company that has just filed a claim for bankruptcy recently. The company now provides you with two alternatives to recover some of your money. 
    \begin{enumerate}[label=(\alph*),nosep,leftmargin=3\parindent] 
        \vspace{0.1cm} 
        \item You will save \$2000 of your money 
        \vspace{0.1cm} 
        \item You will take part in a random drawing procedure with exactly a one-third probability of saving all \$6000 of your money, and two-thirds probability of saving none of your money. 
    \end{enumerate}
\vspace{0.3cm}
Q7: Imagine that in one particular state it is projected that 1000 students will dropout of school during the year, two programs have been prosed to address this problem, but only one can be implemented. Based on other states experiences with programs, estimates of the outcomes that can be expected for each program can be made. 
    \begin{enumerate}[label=(\alph*),nosep,leftmargin=3\parindent] 
        \vspace{0.1cm} 
        \item Program 1: 600 of the 1000 students will drop out of school 
        \vspace{0.1cm} 
        \item Program 2: 2/5 chance that none of the 1000 students will drop out of school and 3/5 chance that all 1000 students will drop out of school 
    \end{enumerate}
\vspace{0.3cm}
Q8: Assume that you have just been given a gift of \$1000. 
    \begin{enumerate}[label=(\alph*),nosep,leftmargin=3\parindent] 
        \vspace{0.1cm} 
        \item Taking an additional \$500 for sure. 
        \vspace{0.1cm} 
        \item Flipping a coin and winning another \$1000 if heads comes up or getting no additional money if tails comes up. 
    \end{enumerate}
\vspace{0.3cm}
Q9(i): Imagine that you face the following pair of concurrent decisions. First examine both decisions, then indicate the options you prefer. 
    \begin{enumerate}[label=(\alph*),nosep,leftmargin=3\parindent] 
        \vspace{0.1cm} 
        \item A sure gain of \$240 
        \vspace{0.1cm} 
        \item 25\% chance to gain \$1000, and 75\% chance to gain nothing 
    \end{enumerate}
\vspace{0.3cm}
Q9(ii): Imagine that you face the following pair of concurrent decisions. First examine both decisions, then indicate the options you prefer. 
    \begin{enumerate}[label=(\alph*),nosep,leftmargin=3\parindent] 
        \vspace{0.1cm} 
        \item A sure loss of \$750 
        \vspace{0.1cm} 
        \item 75\% chance to lose \$1000, and 25\% chance to lose nothing 
    \end{enumerate}
\vspace{0.3cm}
Q9(iii): Imagine that you face the following pair of concurrent decisions. First examine both decisions, then indicate the options you prefer. 
    \begin{enumerate}[label=(\alph*),nosep,leftmargin=3\parindent] 
        \vspace{0.1cm} 
        \item 25\% chance to win \$240, and 75\% chance to lose \$760 
        \vspace{0.1cm} 
        \item 25\% chance to win \$250, and 75\% chance to lose \$750 
    \end{enumerate}
\vspace{0.3cm}
Q10: You are staying in a hotel room on vacation. You paid \$6.95 to see a movie on pay TV. After 5 minutes you are bored and the movie seems pretty bad. Would you continue to watch the movie or not? 
    \begin{enumerate}[label=(\alph*),nosep,leftmargin=3\parindent] 
        \vspace{0.1cm} 
        \item Continue to watch 
        \vspace{0.1cm} 
        \item Turn it off and lose \$6.95 
    \end{enumerate}
\vspace{0.3cm}
Q11: Imagine that your country is preparing for the outbreak of an unusual disease, which is expected to kill 600 people. Two alternative programs to combat the disease have been proposed. Assume that the exact scientific estimate of the consequences of the programs are as follows: 
    \begin{enumerate}[label=(\alph*),nosep,leftmargin=3\parindent] 
        \vspace{0.1cm} 
        \item If Program A is adopted, 200 people will be saved 
        \vspace{0.1cm} 
        \item If Program B is adopted, there is 1/3 probability that 600 people will be saved, and 2/3 probability that no people will be saved 
    \end{enumerate}
\vspace{0.3cm}
Q12: Imagine that you have decided to see a play where admission is \$10 per ticket. As you enter the theatre you discover that you have lost a \$10 bill. 
    \begin{enumerate}[label=(\alph*),nosep,leftmargin=3\parindent] 
        \vspace{0.1cm} 
        \item Still pay \$10 for a ticket for the play 
        \vspace{0.1cm} 
        \item Don't pay \$10 for a ticket for the play 
    \end{enumerate}
\vspace{0.3cm}
Q13: Consider the following two stage game. In the first stage, there is a 75\% chance to end the game without winning anything, and a 25\% chance to move into the second stage. If you reach the second stage, you have a choice between: A sure win of \$30 and 80\% chance to win \$45 
    \begin{enumerate}[label=(\alph*),nosep,leftmargin=3\parindent] 
        \vspace{0.1cm} 
        \item A sure win of \$30  
        \vspace{0.1cm} 
        \item 80\% chance to win \$45 
    \end{enumerate}
\vspace{0.3cm}
Q14: Imagine that six people in your family, including both of your parents, your brothers and your sisters, are infected by a fatal disease. Two alternative medical plans to treat the disease have been proposed. Assume that the exact scientific estimates of the consequences of the plans are as follows:  
    \begin{enumerate}[label=(\alph*),nosep,leftmargin=3\parindent] 
        \vspace{0.1cm} 
        \item If plan A is adopted, two of them will be saved. 
        \vspace{0.1cm} 
        \item If plan B is adopted, there is a one-third probability that all six of them will be saved, and two-thirds probability that none of them will be saved. 
    \end{enumerate}
\vspace{0.3cm}
Q15: Your are presented with the following report from the head of a special team assigned to investigate the prospects of a project in Arizona: Our new analysis indicates that, if we choose to compete with ATC, we would face the possibility of capturing only a small market share. This would give us an after-tax return on investment of as little as 10\%, while capturing a large market share would give us a return of 22\%. We estimate our chance of getting a small market share to be 2 in 3. If we were to team up with ATC on the terms proposed, our return would be 14\% after tax, with the same total investment. 
    \begin{enumerate}[label=(\alph*),nosep,leftmargin=3\parindent] 
        \vspace{0.1cm} 
        \item Compete with ATC: 1/3 chance of gaining a large market share of 22\% and 2/3 chance of gaining a small market share of 10\% 
        \vspace{0.1cm} 
        \item Don't compete with ATC: 100\% chance of capturing 14\% market share 
    \end{enumerate}
\vspace{0.3cm}
Q16: A committee found a fish disease in a nearby lake. About 12 fish species (among them the most popular dining fish) have the Proliferative Kidney Disease (PKD). This is a chronically developing infectious disease which can have deadly consequences for the fish. Young fish are especially susceptible, while others seem to be immune against an infection. Experts suggest that PKD is one cause of declining fish catches. The researchers assume human activities and water pollution foster the spread of the disease. They are considering releasing more fish into the lake to control the epidemic. Imagine that you are a government official of the adjacent village.  
    \begin{enumerate}[label=(\alph*),nosep,leftmargin=3\parindent] 
        \vspace{0.1cm} 
        \item Option A: If the release of fish is implemented, 4 fish species will survive. 
        \vspace{0.1cm} 
        \item Option B: If the release of fish is implemented, there is 1/3 probability that all of the 12 fish species will survive, and 2/3 probability that none of them will survive. 
    \end{enumerate}
\vspace{0.3cm}
Q17: Imagine a refinery that processes petroleum products. An investigation found that due to tank leaks, both soil and drinking water became contaminated. Due to this contamination 720 children from the adjacent village have a fatal disease. There is agreement among experts that children will not suffer health problems, provided they have a strong immune system. Otherwise, it is likely that children will have serious health problems. A vaccine against this disease has been developed and tested. However, the vaccine sometimes can cause side effects that can be fatal too. You are an environmental activist with much influence on the local hospital and you have to decide if you want to lobby for the vaccination or not. 
    \begin{enumerate}[label=(\alph*),nosep,leftmargin=3\parindent] 
        \vspace{0.1cm} 
        \item Option C:  If the vaccination is adopted, the health of 480 children will be damaged for sure. 
        \vspace{0.1cm} 
        \item Option D: If the vaccination is adopted, there is a one-third probability that the health of none of the 720 children will be damaged, and a two-thirds probability that the health of all 720 of them will be damaged. 
    \end{enumerate}
\vspace{0.3cm}
Q18: Assume that you have just been given a gift of \$2000. But you now are forced to choose between the following two alternatives:  
    \begin{enumerate}[label=(\alph*),nosep,leftmargin=3\parindent] 
        \vspace{0.1cm} 
        \item Losing \$500 for sure 
        \vspace{0.1cm} 
        \item Flipping a coin and losing \$1000 if heads comes up or losing nothing if tails comes up 
    \end{enumerate}
\vspace{0.3cm}
Q19: Which of the following options do you prefer? 
    \begin{enumerate}[label=(\alph*),nosep,leftmargin=3\parindent] 
        \vspace{0.1cm} 
        \item 25\% chance to win \$30 
        \vspace{0.1cm} 
        \item 20\% chance to win \$45 
    \end{enumerate}
\vspace{0.3cm}
Q20: Imagine that you are about to purchase a jacket for \$125, and a calculator for \$15. The calculator salesman informs you that the calculator you wish to buy is on sale for \$10 at the other branch of the store, located 20 minutes drive away. 
    \begin{enumerate}[label=(\alph*),nosep,leftmargin=3\parindent] 
        \vspace{0.1cm} 
        \item Make the trip to the other store and save 5 dollars but lose 20 minutes 
        \vspace{0.1cm} 
        \item Don't make the trip to the other store and save 20 minutes but lose 5 dollars 
    \end{enumerate}
\vspace{0.3cm}
Q21: Imagine that six people in your family, including both of your parents, your brothers and your sisters, are infected by a fatal disease. Two alternative medical plans to treat the disease have been proposed. Assume that the exact scientific estimates of the consequences of the plans are as follows: 
    \begin{enumerate}[label=(\alph*),nosep,leftmargin=3\parindent] 
        \vspace{0.1cm} 
        \item If plan A is adopted, four of them will die. 
        \vspace{0.1cm} 
        \item If plan B is adopted, there is a one-third probability that none of them will die, and two-thirds probability that all six of them will die. 
    \end{enumerate}
\vspace{0.3cm}
Q22: Imagine you recieve a letter from the president of a subsidiary describing a dilemma concerning whether to fight an impending patent violation suit or settle out of court that reads: If we do not agree to this proposal, PMG will file their suit. Going to court would involve the possibility of keeping the Duraplast line and incurring only a small sum for legal expenses. If we lose in court, we will incur \$1,100,000 in damages. Our corporate lawyer, Mr. Bell, and our outside law firm agree that we have a 1 in 3 chance of winning the case. 
    \begin{enumerate}[label=(\alph*),nosep,leftmargin=3\parindent] 
        \vspace{0.1cm} 
        \item Agree to the proposal (no lawsuit) 
        \vspace{0.1cm} 
        \item Disagree to the proposal: 1/3 chance of winning the case 
    \end{enumerate}
\vspace{0.3cm}
Q23: Imagine that your country is preparing for the outbreak of an unusual disease, which is expected to kill 600 people. Two alternative programs to combat the disease have been proposed. Assume that the exact scientific estimate of the consequences of the programs are as follows: 
    \begin{enumerate}[label=(\alph*),nosep,leftmargin=3\parindent] 
        \vspace{0.1cm} 
        \item If Program C is adopted 400 people will die.  
        \vspace{0.1cm} 
        \item If Program D is adopted there is 1/3 probability that no one will die, and 2/3 probability that 600 people will die.  
    \end{enumerate}
\vspace{0.3cm}
Q24: You are staying in a hotel room on vacation. You turn on the TV and there is a movie on. After 5 minutes you are bored and the movie seems pretty bad. Would you continue to watch the movie or not? 
    \begin{enumerate}[label=(\alph*),nosep,leftmargin=3\parindent] 
        \vspace{0.1cm} 
        \item Continue to watch 
        \vspace{0.1cm} 
        \item Turn it off 
    \end{enumerate}
\vspace{0.3cm}
Q25: Imagine that six people are infected by a fatal disease. Two alternative medical plans to treat the disease have been proposed. Assume that the exact scientific estimates of the consequences of the plans are as follows:  
    \begin{enumerate}[label=(\alph*),nosep,leftmargin=3\parindent] 
        \vspace{0.1cm} 
        \item If plan A is adopted, four people will die. 
        \vspace{0.1cm} 
        \item If plan B is adopted, there is a one-third probability that none of them will die, and two-thirds probability that all six people will die.  
    \end{enumerate}
\vspace{0.3cm}
Q26: Imagine that 6000 pieces of precious paintings in a world-famous museum are accidentally exposed to a disastrous chemical pollution. Two alternative plans to rescue these art treasures have been proposed. Assume that the exact estimates of the consequences of the plans made by scientists are as follows: 
    \begin{enumerate}[label=(\alph*),nosep,leftmargin=3\parindent] 
        \vspace{0.1cm} 
        \item If plan A is adopted, 2000 pieces will be saved from the chemical pollution. 
        \vspace{0.1cm} 
        \item If plan B is adopted, there is a one-third probability that all the 6000 paintings will be saved, and two-thirds probability that none of these paintings will be saved. 
    \end{enumerate}
\vspace{0.3cm}
Q27: Imagine that in one particular state it is projected that 1000 students will dropout of school during the year, two programs have been prosed to address this problem, but only one can be implemented. Based on other states experiences with programs, estimates of the outcomes that can be expected for each program can be made. 
    \begin{enumerate}[label=(\alph*),nosep,leftmargin=3\parindent] 
        \vspace{0.1cm} 
        \item Program 1: 400 of the 1000 students will stay in school 
        \vspace{0.1cm} 
        \item Program 2: 2/5 chance that all 1000 students will stay in school and 3/5 chance that none of the 1000 will stay in school 
    \end{enumerate}
\vspace{0.3cm}
Q28: Imagine that you have lung cancer and you must choose between two therapies: surgery and radiation. Surgery for lung cancer involves an operation on the lungs. Most patients are in the hospital for two or three weeks and have some pain around their incisions; they spend a month or so recuperating at home. After that, they generally feel fine. Radiation therapy for lung cancer involves the use of radiation to kill the tumor and requires coming to the hospital about four times a week for six weeks. Each treatment takes a few minutes and during the treatment, patients lie on a table as if they were having an x-ray. During the course of the treatment, some patients develop nausea and vomiting, but by the end of the six weeks they also generally feel fine. Thus, after the initial six or so weeks, patients treated with either surgery or radiation therapy feel about the same. 
    \begin{enumerate}[label=(\alph*),nosep,leftmargin=3\parindent] 
        \vspace{0.1cm} 
        \item Surgery: Of 100 people having surgery, 10 die during surgery or the postoperative period, 32 die by the end of one year and 66 die by the end of five years. 
        \vspace{0.1cm} 
        \item Radiation Therapy: Of 100 people having radiation therapy, none die during treatment, 23 die by the end of one year and 78 die by the end of five years. 
    \end{enumerate}
\vspace{0.3cm}
Q29: Imagine that you have decided to see a play and paid the admission price of \$10 per ticket. As you enter the theatre you discover that you have lost the ticket. The seat was not marked and the ticket cannot be recovered. 
    \begin{enumerate}[label=(\alph*),nosep,leftmargin=3\parindent] 
        \vspace{0.1cm} 
        \item Pay \$10 for another ticket 
        \vspace{0.1cm} 
        \item Don't pay \$10 for another ticket 
    \end{enumerate}
\vspace{0.3cm}
Q30: Your are presented with the following report from the head of a special team assigned to investigate the prospects of a project in Arizona: Our new analysis indicates that, if we choose to compete with ATC, we would have the possibility of capturing a large market share. This would give us an after-tax return on investment of as much as 22\%, while capturing a small market share would give us a return of only 10\%. We estimate a 1 in 3 chance of getting a large market share. If we were to team up with ATC on the terms proposed, our return would be 14\% after tax, with the same total investment. 
    \begin{enumerate}[label=(\alph*),nosep,leftmargin=3\parindent] 
        \vspace{0.1cm} 
        \item Compete with ATC 
        \vspace{0.1cm} 
        \item Don't compete with ATC 
    \end{enumerate}
\vspace{0.3cm}
Q31: A large car manufacturer has recently been hit with a number of economic difficulties and it appears as if three plants need to be closed and 6000 employees laid off. The vice-president of production has been exploring alternative ways to avoid this crisis and has developed two plans: 
    \begin{enumerate}[label=(\alph*),nosep,leftmargin=3\parindent] 
        \vspace{0.1cm} 
        \item Plan A: This plan will save 1 plant and 2000 jobs 
        \vspace{0.1cm} 
        \item Plan B: : This plan has a 1/3 probability of saving all 3 plants and all 6000 jobs, but has a 2/3 probability of saving no plants and no jobs 
    \end{enumerate}
\vspace{0.3cm}
Q32: Imagine that you are about to purchase a jacket for \$15, and a calculator for \$125. The calculator salesman informs you that the calculator you wish to buy is on sale for \$120 at the other branch of the store, located 20 minutes drive away. 
    \begin{enumerate}[label=(\alph*),nosep,leftmargin=3\parindent] 
        \vspace{0.1cm} 
        \item Make the trip to the other store and save 5 dollars but lose 20 minutes 
        \vspace{0.1cm} 
        \item Don't make the trip to the other store and save 20 minutes but lose 5 dollars 
    \end{enumerate}
\vspace{0.3cm}
Q33: A committee found a fish disease in a nearby lake. About 12 fish species (among them the most popular dining fish) have the Proliferative Kidney Disease (PKD). This is a chronically developing infectious disease which can have deadly consequences for the fish. Young fish are especially susceptible, while others seem to be immune against an infection. Experts suggest that PKD is one cause of declining fish catches. The researchers assume human activities and water pollution foster the spread of the disease. They are considering releasing more fish into the lake to control the epidemic. Imagine that you are a government official of the adjacent village. 
    \begin{enumerate}[label=(\alph*),nosep,leftmargin=3\parindent] 
        \vspace{0.1cm} 
        \item Option C:  If the release of fish is implemented, 8 fish species will die. 
        \vspace{0.1cm} 
        \item Option D: If the release of fish is implemented, there is 2/3 probability that none of the 12 fish species will die, and 1/3 probability that all of the 12 fish species will die. 
    \end{enumerate}
\vspace{0.3cm}
Q34: Imagine that you brought \$6000 worth of stock from a company that has just filed a claim for bankruptcy recently. The company now provides you with two alternatives to recover some of your money. 
    \begin{enumerate}[label=(\alph*),nosep,leftmargin=3\parindent] 
        \vspace{0.1cm} 
        \item You will lose \$4000 of your money 
        \vspace{0.1cm} 
        \item You will take part in a random drawing procedure with exactly a two-thirds probability of losing \$6000 all of your money, and one-third probability of not losing any of your money 
    \end{enumerate}
\vspace{0.3cm}
Q35: Imagine a refinery that processes petroleum products. An investigation found that due to tank leaks, both soil and drinking water became contaminated. Due to this contamination 720 children from the adjacent village have a fatal disease. There is agreement among experts that children will not suffer health problems, provided they have a strong immune system. Otherwise, it is likely that children will have serious health problems. A vaccine against this disease has been developed and tested. However, the vaccine sometimes can cause side effects that can be fatal too. You are an environmental activist with much influence on the local hospital and you have to decide if you want to lobby for the vaccination or not. 
    \begin{enumerate}[label=(\alph*),nosep,leftmargin=3\parindent] 
        \vspace{0.1cm} 
        \item Option A:  If the vaccination is adopted, the health of 240 children will be saved for sure. 
        \vspace{0.1cm} 
        \item Option B: If the vaccination is adopted, there is a one-third probability that the health of all of the 720 children will be saved, and a two-thirds probability that the health of none of them will be saved. 
    \end{enumerate}
\vspace{0.3cm}
Q36: Imagine that six people are infected by a fatal disease. Two alternative medical plans to treat the disease have been proposed. Assume that the exact scientific estimates of the consequences of the plans are as follows:  
    \begin{enumerate}[label=(\alph*),nosep,leftmargin=3\parindent] 
        \vspace{0.1cm} 
        \item If plan A is adopted, two people will be saved. 
        \vspace{0.1cm} 
        \item If plan B is adopted, there is a one-third probability that all six people will be saved, and two-thirds probability that none of them will be saved.  
    \end{enumerate}

\subsection{Group Level Results}

Table \ref{appendix:table} shows the full table corresponding to table \ref{tab:micro} in the paper with extra information regarding the choices made in each human experiment. $P$ represents the percentage of people that chose the risky choice given either the gain frame or loss frames, the true $P_{gain}$ and $P_{loss}$ values, and $n$ represents the number of human subjects who answered each frame which corresponds to the number of decisions our model made to produce results in the table.

\newpage
\onecolumn

{ \fontsize{7.25}{7.25}\selectfont 
 \begin{longtable*}[b]{c c c c c c | c c c | c c c }     \hline 
 & \multicolumn{2}{c}{Gain Frame}  &  \multicolumn{2}{c}{Loss Frame} &  & \multicolumn{3}{c}{BR Baseline} & \multicolumn{3}{|c}{CDM Model} \\
 Reference & n  & P  & n  & P & Actual LOR & LOR & SE &  $\chi^2$ & LOR & SE &  $\chi^2$\\
 \hline 
 \multicolumn{12}{c}{Standard ADP; one presentation, between-subjects, low PISA} \\
 \citet{tversky1981framing} & \textbf{152} & \textbf{28} & \textbf{155} & \textbf{78} & \textbf{2.20} & \textbf{1.65*} & \textbf{.26} & \textbf{4.34} & \textbf{1.86} & \textbf{.26} & \textbf{1.83} \\

 \citet{reyna1991fuzzy} & 36 & 53 & 36 & 81 & 1.31 & 1.72 & .54 & .57 & 1.79 & .54 & .79 \\

 \citet{tindale1993framing} & 144 & 42 & 144 & 79 & 1.63 & 1.71 & .26 & .10 & 1.77 & .26 & .27 \\

 \citet{wang1995perceived} & 50 & 40 & 50 & 68 & 1.16 & 1.73 & .42 & 1.83 & 1.74 & .45 & 1.71 \\

 \citet{highhouse1996perspectives} & 122 & 29 & 122 & 74 & 1.94 & 1.68 & .29 & .82 & 1.69 & .28 & .80  \\

 \citet{wang1996framing} & 31 & 42 & 34 & 77 & 1.50 & 1.71 & .54 & .14 & 1.62 & .55 & .05 \\

 \citet{Stanovich1998} & 148 & 32 & 144 & 65 & 1.37 & 1.74 & .25 & 2.34 & 1.78 & .26 & 2.43  \\

 \citet{druckman2001evaluating} & 50 & 32 & 55 & 77 & 1.93 & 1.70 & .44 & .27 & 1.51 & .43 & .94 \\

 \citet{druckman2001using} & 69 & 32 & 79 & 76 & 1.91 & 1.70 & .37 & .34 & 1.66 & .36 & .49 \\
 
 \citet{mayhorn2002decisions}, Young & 29 & 24 & 29 & 86 & 2.98 & 1.68 & .69 & 3.52 & 1.50 & .58 & 5.89* \\
 
 \citet{mayhorn2002decisions}, Older & 29 & 21 & 29 & 69 & 2.14 & 1.70 & .61 & .53 & 1.54 & .58 & 1.06 \\

 \citet{LeBoeuf2003}, Exp \#1  & 48 & 49 & 55 & 56 & 1.40 & 1.74  & .25 & 1.77 & 1.54 & .43 & .12 \\

 \citet{LeBoeuf2003}, Exp \#2 & 147 & 25 & 146 & 57 & 1.47 & 1.71  & .43 & .32 & 1.58 & .25 & .13  \\

 \cite{stein2012framing} & 47 & 40 & 57 & 68 & 1.16 & 1.73 & .41 & 1.89 & 1.72 & .44 & 1.55  \\
 \multicolumn{12}{r}{TOTAL of 14 predicted: 13 (93\%)} \\
 \hline 
 \multicolumn{12}{c}{Standard ADP; one presentation, between-subjects, high PISA} \\
 \citet{takemura1994influence} & 45 & 20 & 45 & 69 & 2.18 & 1.39 & .49 & 2.56 & 1.52 & .47 & 2.02 \\

 \citet{mandel2001gain} & 26 & 54 & 26 & 85 & 1.55 & 1.44 & .67 & .03 & 1.40 & .61 & .08 \\

 \citet{fischer2008selective} & 17 & 36 & 17 & 77 & 1.78 & 1.43 & .76 & .21 & 1.43 & .78 & .20  \\

 \citet{zhang2008social} \#1 & 65 & 66 & 68 & 87 & 1.21 & 1.47 & .44 & .34 & 1.44 & .37 & .29 \\

 \citet{zhang2008social} \#2 & 45 & 67 & 48 & 88 & 1.25 & 1.46 & .54 & .14 & 
 1.41 & .45 & .08 \\

 \citet{zhang2008effect}, Military & 134 & 54 & 130 & 83 & 1.44 & 1.44 & .29 & .00 & 1.48 & .27 & .05 \\

 \citet{zhang2008effect}, Civilian & 60 & 65 & 58 & 90 & 1.54 & 1.43 & .51 & .04 & 1.48 & .40 & .05 \\

 \citet{haerem2011military} & 29 & 59 & 26 & 73 & .65 & 1.48 & .58 & 2.02 & 1.46 & .59 & 1.93\\

 \citet{okder2012illusion} & 52 & 37 & 53 & 76 & 1.68 & 1.42 & .43 & .34 & 1.66 & .43 & .01 \\

 \citet{kuhberger2013choice}, Exp \#1 & 63 & 32 & 63 & 68 & 1.53 & 1.43 & .38 & .06 & 1.45 & .39 & .02 \\

 \citet{kuhberger2013choice}, Exp \#2 & 14 & 57 & 15 & 73 & .72 & 1.46 & .80  & .85 & 1.52 & .86 & .87 \\

 \citet{mandel2014framing}, Exp \#2 & 38 & 42 & 38 & 74 & 1.35 & 1.45 & .49 & .04  & 1.57 & .51 & .16 \\

 \citet{mandel2014framing}, Exp \#3 & 25 & 32 & 25 & 80 & 2.14 & 1.42 & .66 & 1.20 & 1.54 & .63 & .92 \\
 \multicolumn{12}{r}{TOTAL of 13 predicted: 13 (100\%)} \\
 \hline 
 \multicolumn{12}{c}{Standard ADP; within-subjects, low PISA} \\
 \citet{Stanovich1998} & 292 & 32 & 292 & 54 & .9 & .94 & .24 & 1.58 & .96 & .17 & .08 \\
 \citet{Levin2002} & 102 & 28 & 102 & 56 & 1.2 & .92 & .30 & .94 & .94 & .29 & .74 \\
 \citet{LeBoeuf2003} Exp \#2 & \textbf{287} & \textbf{25} & \textbf{287} & \textbf{46} & \textbf{.57} & \textbf{1.05*} & \textbf{.17} & \textbf{7.86} & \textbf{.81} & \textbf{.17} & \textbf{.52} \\
 \multicolumn{12}{r}{TOTAL of 3 predicted: 3 (100\%)} \\
 \hline 
 \multicolumn{12}{c}{Standard ADP; multiple presentations, between-subjects, low PISA} \\
 \citet{Fagley1990}, Exp \#1 & 94 & 51 & 96 & 70 & .79 & .95 & .30 & .27 & 1.38 & .31 & 3.01  \\
 \citet{Fagley1990}, Exp \#2 & 54 & 39 & 55 & 73 & 1.43 & .92 & .41 & 1.55 & 1.28 & .41 & .16  \\
 \citet{Miller1991} & 23 & 43 & 23 & 67 & .89 & .94 & .61 & .01 & 1.40 & .65 & .40 \\
 \citet{Jou1996} & 80 & 35 & 80 & 80 & 2.01 &  .87* & .36 & 9.66 & 1.13 & .33 & 6.99*  \\
 \citet{Ronnlund2005} Young & 32 & 41 & 32 & 69 & 1.19 & .93 & .53 &.23 & 1.30 & .55 & .07  \\
 \citet{Ronnlund2005},
 Older & 32 & 28 & 32 & 56 & 1.17 & .94 & .52 & .20 & 1.40 & .55 & .16  \\
 \multicolumn{12}{r}{TOTAL of 6 predicted: 5 (83\%)} \\
 \hline 
 \multicolumn{12}{c}{Standard ADP; multiple presentations, between-subjects, High PISA} \\
 \citet{KUHBERGER1995230}, Exp \#1 & 25 & 48 & 23 & 78 & 1.36 & .73 & .64 & 1.31 & .86 & .61 & .64  \\
 \citet{KUHBERGER1995230}, Exp \#2 & 16 & 56 & 14 & 57 & .04 & .81 & .74 & .81 & 1.10 & .80 & 1.76  \\
 \citet{Druckman2008} & 101 & 45 & 113 & 67 & .94 & .71 & .28 & 1.84 & .46 & .28 & 2.57  \\
  \multicolumn{12}{r}{TOTAL of 3 predicted: 3 (100\%)} \\
  \hline 
 \multicolumn{12}{c}{Allais Paradox gambles; low PISA} \\
\citet{conlisk1989three} & 236 & 49 & 236 & 86 & 1.83 & 1.68 & .23 & .44 & 1.70 & .21 & .55 \\
\citet{carlin1990allais} & 65 & 40 & 65 & 78 & 1.7 & 1.71 & .39 & 0 & 1.75 & .45 & .03\\
 \multicolumn{12}{r}{TOTAL of 2 predicted: 2 (100\%)} \\
  \hline 
 \multicolumn{12}{c}{Allais Paradox gambles; low PISA} \\
 \citet{huck2012allais} Laboratory & 70 & 66 & 70 & 87 & 1.26 & 1.73 & .44 & 1.14 & 1.32 & .46 & .03 \\
  \multicolumn{12}{r}{TOTAL of 1 predicted: 1 (75\%)} \\
  \hline 
 \multicolumn{12}{c}{Other framing problems; multiple presentations, between-subjects, low PISA} \\
\citet{reyna2014developmental} College Students  & 63 & 35 & 63 & 55 & .85 & .95 & .37 & .02 & .99 & .37 & .21\\
\citet{reyna2014developmental} Adults  & 54 & 40 & 54 & 60 & .8 & .95 & .39 & .01 & 1.12 & .41 & .60 \\
\citet{reyna2014developmental} Experts  & 36 & 38 & 36 & 71 & 1.37 & .93 & .5 & .18 & .69 & .49 & 2.01 \\
 \multicolumn{12}{r}{TOTAL of 3 predicted: 3 (100\%)} \\
    \hline 
 \multicolumn{12}{c}{Other framing problems; multiple presentations, between-subjects, high PISA} \\
\citet{kuhberger1995framing} Plant \#1  & 25 & 52 & 23 & 83 & 1.48 & .73 & .68 & 1.61 & .97 & .61 & .79 \\
\citet{kuhberger1995framing} Cancer \#1 & 24 & 38 & 25 & 48 & .43 & .8 & .58 & .2 & .21 & .60 & .11 \\
\citet{kuhberger1995framing} Plant \#2 & 16 & 19 & 17 & 71 & 2.34 & .7 & .83 & 4.43 & .34 & .73 & 7.50*\\
\citet{kuhberger1995framing} Cancer \#2 & 16 & 69 & 14 & 64 & -.2 & .82 & .78 & 1.39 & .97 & .82 & 2.11\\
 \multicolumn{12}{r}{TOTAL of 4 predicted: 3 (75\%)} \\
    \hline 
 \multicolumn{12}{c}{Other framing problems; multiple presentations, between-subjects, mixed PISA} \\
\citet{kuhberger2010risky} Water contamination & 93 & 33 & 93 & 73 & 1.69 & 1.27 & .32 & 1.78 &1 .26 & .31 & 2.01 \\
\citet{kuhberger2010risky} Crops & 93 & 33 & 93 & 59 & 1.06 & 1.36 & .3 & .92 & 1.32 & .31 & .64 \\
\citet{kuhberger2010risky} Fish disease & 93 & 28 & 93 & 59 & 1.32 & 1.32 & .31 & 0 & 1.20 & .31 & .12 \\
\citet{kuhberger2010risky} Endangered forest & 93 & 24 & 93 & 55 & 1.37 & 1.31 & .32 & .03 & 1.23 & .31 & .16 \\
 \multicolumn{12}{r}{TOTAL of 4 predicted: 4 (100\%)} \\
     \hline 
 \multicolumn{12}{c}{Zero-complement truncated framing problems, one presentation; framing manipulated between-subjects} \\
\citet{reyna1991fuzzy}  & 35 & 51 & 36 & 58 & .28 & 0 & .48 & .34 & .41 & .49 & .07 \\
\citet{mandel2001gain} Exp\#1 & 23 & 48 & 25 & 72 & 1.03 & 0 & .61 & 2.86 & .30 & .60 & 1.47 \\
\citet{mandel2001gain} Exp\#2  & 36 & 64 & 38 & 63 & -.03 & 0 & .48 & 0 & .35 & .48 & .70 \\
 \multicolumn{12}{r}{TOTAL of 3 predicted: 3 (100\%)} \\
     \hline 
 \multicolumn{12}{c}{Zero-complement truncated framing problems; multiple presentations, framing manipulated between-subjects} \\
\citet{kuhberger2010risky} Water contamination   & 93 & 54 & 93 & 65 & .45 & 0 & .3 & 2.21 & 0 & .3 & 2.21 \\
\citet{kuhberger2010risky} Crops & 93 & 54 & 93 & 43 & -.43 & 0 & .3 & 2.14 & .34 & .30 & 6.83* \\
\citet{kuhberger2010risky} Fish disease & 93 & 63 & 93 & 43 & -.83 & 0 & .3 & 7.68* & .37 & .30 & 15.74* \\
\citet{kuhberger2010risky} Endangered forest & 93 & 40 & 93 & 43 & .13 & 0 & .3 & .2 & .23 & .30 & .13 \\
\citet{reyna2014developmental} College Students & 63 & 43 & 63 & 49 & .25 & 0 & .36 & .47 & .26 & .36 & 0 \\
\citet{reyna2014developmental} Adults  & 54 & 51 & 54 & 55 & .18 & 0 & .39 & .23 & .29 & .39 & .75 \\
\citet{reyna2014developmental} Experts & 36 & 52 & 36 & 62 & .41 & 0 & .48 & .74 & .20 & .48 & .19 \\
 \multicolumn{12}{r}{TOTAL of 7 predicted: 5 (71.4\%)} \\
     \hline 
 \multicolumn{12}{c}{Nonzero-complement truncated framing problems; one presentation, between-subjects, low PISA} \\
 \citet{reyna1991fuzzy} & 35 & 26 & 37 & 81 & 2.52 & 3.44 & .57 & 2.59 & 2.29 & .57 & .13  \\
  \multicolumn{12}{r}{TOTAL of 1 predicted: 1 (100\%)} \\
     \hline 
 \multicolumn{12}{c}{Nonzero-complement truncated framing problems; multiple presentations, between-subjects, low PISA} \\
\citet{reyna2014developmental} College Students & 63 & 23 & 63 & 60 & 1.61 & 1.9 & .4 & .51 & 2.19 & .42 & 1.90  \\
\citet{reyna2014developmental} Adults  & 54 & 26 & 54 & 73 & 2.05 & 1.87 & .44 & .15 & 1.83 & .44 & 0.23 \\
\citet{reyna2014developmental} Experts & 36 & 20 & 36 & 81 & 2.84 & 1.85 & .59 & 2.72 & 1.86 & .54 & 3.29 \\
 \multicolumn{12}{r}{TOTAL of 3 predicted: 3 (100\%)} \\
     \hline 
 \multicolumn{12}{c}{Nonzero-complement truncated framing problems; one presentation, between-subjects, low PISA} \\
\citet{kuhberger2010risky} Water contamination & 93 & 25 & 93 & 85 & 2.84 & 2.61 & .38 & .39 & 2.31 & .35 & 2.24 \\
\citet{kuhberger2010risky} Crops & 93 & 29 & 93 & 78 & 2.19 & 2.7 & .34 & 2.28 & 2.27 & .35 & .09 \\
\citet{kuhberger2010risky} Fish diseas & 93 & 22 & 93 & 83 & 2.87 & 2.61 & .37 & .49 & 2.21 & .34 & 3.48 \\
\citet{kuhberger2010risky} Endangered forest & 93 & 15 & 93 & 65 & 2.33 & 2.68 & .36 & .96 & 2.30 & .35 & .02 \\
 \multicolumn{12}{r}{TOTAL of 4 predicted: 4 (100\%)} \\
     \hline 
 \multicolumn{12}{c}{Certain-option disambiguated problems; single presentation, between-subjects} \\
\citet{mandel2001gain} Exp \#1  & 23 & 52 & 22 & 50 & -.09 & 0 & .6 & .02 & .28 & .62 & .34 \\
\citet{mandel2014framing} Exp \#3 & 22 & 41 & 24 & 50 & .37 & 0 & .6 & .38 & .03 & .60 & .31 \\
 \multicolumn{12}{r}{TOTAL of 2 predicted: 2 (100\%)} \\
     \hline 
 \multicolumn{12}{c}{Certain-option disambiguated problems; multiple presentations, between-subjects} \\
\citet{kuhberger1995framing} ADP \#1 & 26 & 62 & 23 & 57 & -.21 & 0 & .58 & .13 & .18 & .59 & .45 \\
\citet{kuhberger1995framing} Plant \#1 & 26 & 46 & 23 & 52 & .24 & 0 & .57 & .18 & .11 & .59 & .05\\
\citet{kuhberger1995framing} Cancer \#1 & 24 & 50 & 23 & 35 & -.63 & 0 & .6 & 1.1 & .35 & .60 & 2.61 \\
\citet{kuhberger1995framing} ADP \#2 & 22 & 41 & 19 & 37 & -.17 & 0 & .64 & .07 & .28 & .65 & .47 \\
\citet{kuhberger1995framing} Plant \#2 & 13 & 31 & 19 & 37 & .27 & 0 & .77 & .13 & .17 & .75 & .02 \\
\citet{kuhberger1995framing} Cancer \#2 & 7 & 71 & 13 & 62 & -.45 & 0 & 1.01 & .19 & .13 & 1.02 & .28 \\
 \multicolumn{12}{r}{TOTAL of 6 predicted: 6 (100\%)} \\
     \hline 
 \multicolumn{12}{c}{“400 not saved” certain-option disambiguated and truncated problems; multiple presentations, between-subjects, high
PISA} \\
\citet{kuhberger1995framing} ADP \#1 & 25 & 60 & 23 & 43 & -.67 & -.79 & .59 & .04 & .07 & .59 & 1.64\\
\citet{kuhberger1995framing} Plant \#1 & 27 & 44 & 23 & 57 & .49 & -.88 & .57 & 5.68* & .11 & .58 & .50 \\
\citet{kuhberger1995framing} Cancer \#1 & 24 & 75 & 23 & 43 & -1.36 & -.74 & .63 & .98 & .11 & .60 & 6.21* \\
\citet{kuhberger1995framing} ADP \#2 & 16 & 50 & 19 & 37 & -.54 & -.79 & .69 & .13 & .12 & .70 & .85 \\
\citet{kuhberger1995framing} Plant \#2 & 14 & 57 & 14 & 50 & -.29 & -.8 & .76 & .45 & .15 & .79 & .29\\
\citet{kuhberger1995framing} Cancer \#2 & 14 & 50 & 16 & 44 & -.25 & -.8 & .73 & .56 & .15 & .76 & .26 \\
 \multicolumn{12}{r}{TOTAL of 5 predicted: 6 (83.3\%)} \\
     \hline 
 \multicolumn{12}{c}{Certain-option disambiguated, zero-complement truncated problems; single presentation, between-subjects, high PISA} \\
 \citet{mandel2014framing}, Exp \#3  & 26 & 58 & 25 & 32 & -1.06 & -1.46 & .58 & .45 & .02 & .58 & 3.59 \\
 \multicolumn{12}{r}{TOTAL of 1 predicted: 1 (100\%)} \\
     \hline 
 \multicolumn{12}{c}{“400 not saved vs. 2/3 chance that 600 not saved” truncation problem; single presentation, framing manipulated
between-subjects} \\
\citet{mandel2001gain} Exp \#1 & 23 & 57 & 24 & 58 & .07 & 0 & .59 & .02 & 1.09 & .64 & .01 \\
\citet{mandel2001gain} Exp \#2 & 36 & 64 & 37 & 59 & -.19 & 0 & .48 & .15 & 1.02 & .63 & .00 \\

 \multicolumn{12}{r}{TOTAL of 3 predicted: 3 (100\%)} \\
  \hline 
 \multicolumn{12}{r}{OVERALL TOTAL of 88 predicted: 82 (93.2\%)} \\
 \hline 
 \multicolumn{12}{l}{* Indicates results with Wald statistics over 1 degree of freedom} \\
 \multicolumn{12}{l}{Note: Bolded rows indicate results where CDM outperforms BR} \\
 \label{appendix:table}
 \end{longtable*} 
}

\end{document}